\documentclass[journal]{IEEEtran}
\usepackage{cite}
\usepackage{amsmath,amssymb,amsfonts,bm, bbm}
\usepackage{graphicx,hhline,ragged2e}
\usepackage{algorithm}
\usepackage{algcompatible}
\usepackage{booktabs}
\usepackage{tabularx}
\usepackage[T1]{fontenc}
\usepackage{textcomp,hyperref,adjustbox}

\DeclareMathOperator*{\argmin}{arg\,min}
\usepackage{xcolor}
\usepackage{array}

\begin{document}

\title{Continual Robot Learning using Self-Supervised Task Inference}

\author{Muhammad Burhan Hafez and Stefan Wermter
\thanks{The authors are with the Knowledge Technology Group, Department of Informatics, University of Hamburg, Hamburg, Germany $(${\tt\small\{hafez, wermter\}@uni-hamburg.de}$)$.}}

\maketitle

\begin{abstract}
Endowing robots with the human ability to learn a growing set of skills over the course of a lifetime as opposed to mastering single tasks is an open problem in robot learning. While multi-task learning approaches have been proposed to address this problem, they pay little attention to task inference. In order to continually learn new tasks, the robot first needs to infer the task at hand without requiring predefined task representations. In this paper, we propose a self-supervised task inference approach. Our approach learns action and intention embeddings from self-organization of the observed movement and effect parts of unlabeled demonstrations and a higher-level behavior embedding from self-organization of the joint action-intention embeddings. We construct a behavior-matching self-supervised learning objective to train a novel Task Inference Network (TINet) to map an unlabeled demonstration to its nearest behavior embedding, which we use as the task representation. A multi-task policy is built on top of the TINet and trained with reinforcement learning to optimize performance over tasks. We evaluate our approach in the fixed-set and continual multi-task learning settings with a humanoid robot and compare it to different multi-task learning baselines. The results show that our approach outperforms the other baselines, with the difference being more pronounced in the challenging continual learning setting, and can infer tasks from incomplete demonstrations. Our approach is also shown to generalize to unseen tasks based on a single demonstration in one-shot task generalization experiments.



\end{abstract}


\section{Introduction}
\IEEEPARstart{T}{eaching} robots to perform tasks with minimal knowledge about the environment and without manually programming the desired behavior has always been the driving motivation for the research on robot learning. The field has witnessed remarkable progress over the past decade on a variety of difficult tasks, including manipulation \cite{levine2016end, akkaya2019solving}, locomotion \cite{haarnoja2019learning}, and navigation \cite{zhu2017target}. However, the learning is more oriented towards solving single tasks. To enable multi-task learning, approaches based on knowledge distillation \cite{rusu2016policy, teh2017distral, watkins2021teachable}, meta-learning \cite{NEURIPS2019_d324a0cc, zhou2019watch}, and language conditioning \cite{Lynch-RSS-21, silva2021lancon} have been proposed.  

While these approaches are becoming widely adopted to overcome the limitation of task-specific learning, they have two notable deficiencies. First, they are not compatible with continual learning as they assume a fixed task distribution and treat newly introduced tasks as independent learning problems. Specifically, when a policy to perform a given set of sensorimotor tasks is learned, a complete retraining is often required if a new task is introduced. This is in sharp contrast to the human ability to learn a growing repertoire of skills over the course of a lifetime. Second, they lack an efficient task inference mechanism. This issue is either ignored by using predefined task labels or fixed task representations (e.g., pretrained natural language embeddings) as input or poorly addressed by requiring extensive exploration to gather experience data sufficient to infer a posterior over a latent task variable \cite{rakelly2019efficient, zhang2021metacure}. Humans, on the other hand, only need to observe a demonstration of the desired behavior to successfully infer the task at hand due to their ability to understand and imitate the goal of the observed behavior, not the precise actions \cite{williamson2006precision, rizzolatti2010functional}. 

\begin{figure}[t]
	\centering
		\includegraphics[width=\linewidth]{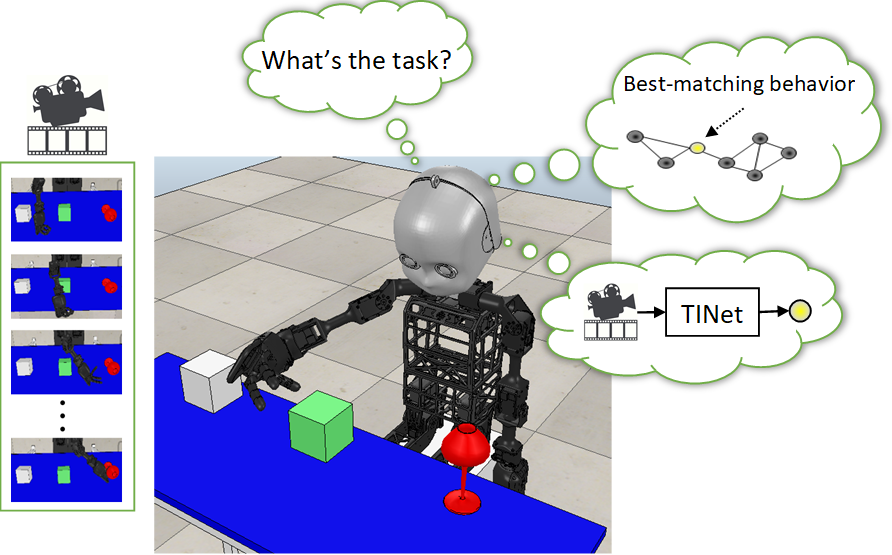}
		\caption{Given an unlabeled demonstration (left), the robot learns to infer the task at hand (e.g., \textit{Grasp the red glass}) by finding the best-matching behavior in the growing, self-organizing network of behaviors and training the proposed task inference network (TINet) to map the demonstration to its best-matching behavior, which we use as the task representation.}
		\label{fig:0}
\end{figure}

Existing multi-task learning approaches that incorporate expert demonstrations can be categorized, according to the way the demonstrations are used, into three distinct groups: i) conditioning the policy on a demonstration embedding \cite{hafez2021behavior, jang2022bc}, ii) training the policy to match demonstration actions with behavior cloning \cite{rahmatizadeh2018vision, lynch2020learning}, and iii) generating a reward based on how close the observed image is to the corresponding one from visual demonstrations \cite{liu2018imitation, kalashnikov2021mt}. A common assumption in these groups of approaches is that the demonstrations are complete. This is a strong assumption in practice for several reasons such as the misalignment between the initial state of the demonstration and that of the robot's environment, sensory noise, self- and object-occlusion, and motion blur. Consider the following example as motivation: If a demonstration of some desired visual manipulation task has missing or corrupted parts (e.g., irretrievable image observations) for any of the above-mentioned reasons, then the control policy can only learn to match the observed actions of the intact part of the demonstration without the ability to recover the remaining actions. In the case of using this demonstration to infer or identify the desired task before solving it by conditioning the policy on the demonstration, the policy will be unable to infer the correct task, and much worse, using this as a training example will impair the learned policy. A multi-task learning approach that enables task inference from incomplete demonstrations is thus needed to relax this assumption. Moreover, such an approach would be consistent with a large body of behavioral and neurocognitive evidence indicating that human children can imitate incomplete task demonstrations \cite{meltzoff1995understanding, tomasello2005understanding, huang2002infants, elsner2007infants, schonebeck2019erps, nielsen200912}. Besides assuming complete demonstrations, the above groups treat each demonstration as one single unit of information, overlooking the fact that a demonstration is a combination of action, which is the observed movement, and intention, which is the observed effect. In contrast, learning movement-effect associations by observation has been found to play an essential role in the developmental changes in human goal-directed imitation, including the ability to imitate behaviors from incomplete demonstrations \cite{elsner2007infants, paulus2013neurocognitive, cooper2013associative}.

In this paper, we propose a multi-task robot learning approach that alleviates the two deficiencies identified earlier—namely, the incompatibility of the existing approaches with continual learning of sensorimotor tasks and the lack of efficient task inference mechanisms. Our approach learns, in an unsupervised manner, a behavior embedding space from unlabelled task demonstrations. We construct a behavior-matching self-supervised learning objective for training a novel Task Inference Network (TINet) to map a given demonstration to its nearest behavior embedding, which we use as the task representation (see Fig. \ref{fig:0}). A multi-task policy is built on top of the task inference network and trained with reinforcement learning to optimize performance over tasks.

In a previous work \cite{hafez2021behavior}, incremental self-organization of visual demonstrations of behaviors was proposed to build a behavior embedding space for efficient task inference in continual robot learning. However, a single self-organizing network was used to learn to map an unlabeled demonstration to a behavior embedding. This means that the network will map an incomplete demonstration to a behavior embedding different from the one that best matches the complete demonstration. Furthermore, a node representing this undesired behavior will be added to the network if the node insertion criterion is met, which impairs the behavior embedding space. In contrast, using two networks separately learning actions and intentions and another learning action-intention associations facilitates finding a correct behavior because from the intention embedding of an incomplete demonstration it will be possible to retrieve a behavior that has the same or similar intention as the one that best matches the complete demonstration. We improve on \cite{hafez2021behavior} by treating the visual demonstration as a combination of an observed movement and an observed effect and learning two separate embeddings for each of these two components of a demonstration, which we call action and intention embeddings respectively. The behavior embedding space is learned by incrementally self-organizing the combined action and intention embeddings. Unlike \cite{hafez2021behavior}, our approach can perform task inference from incomplete demonstrations. This is achieved by randomly sampling a sub-trajectory from the demonstrated trajectory and training the proposed TINet to map both trajectories to the behavior embedding that best matches the demonstrated trajectory with a behavior-matching self-supervised learning objective. Furthermore, the whole learning architecture, including the policy network and the task inference network, is trained end-to-end which allows the task representations to capture the structure of the task at hand.

The primary contributions of our work are summarized as follows:
\begin{itemize}
   \item We develop a hierarchical architecture to learn unsupervised embeddings of actions, intentions, and the resulting action-intention associations from unlabeled demonstration data.
   \item A behavior-matching self-supervised learning objective is proposed to train a task inference network to map an input demonstration to the best matching behavior in the unsupervisedly learned behavior embedding space.
   \item We introduce an end-to-end continual robot learning approach that learns novel tasks over time and can infer tasks from incomplete visual demonstrations.
   \item We evaluate our approach in multi-task learning experiments under the continual learning setting with a humanoid robot and compare it to different multi-task learning baselines.
 \end{itemize}

\section{Related Work}

\subsection{Task-Agnostic Models and Skills}\label{sec:tat-review} 
It has been shown that a world model learned by unsupervised exploration can be used to efficiently solve multiple tasks \cite{sekar2020planning}. First, a task-agnostic exploration policy is trained to collect experience that improves the world model by maximizing expected novelty of future states. A task policy is then trained by imagination inside the world model. The method is found to achieve better zero-shot task performance than other unsupervised methods on continuous control tasks and few-shot performance comparable to a supervised oracle that receives task rewards during exploration. However, fast adaptation to a downstream task requires the world model to be trained in parts of the environment relevant to the task, which cannot be guaranteed with the proposed exploration method. Similar to \cite{sekar2020planning}, Sharma et al.\cite{Sharma-RSS-20} train a reinforcement learning agent without reward supervision and use it to solve downstream tasks. However, the training objective is not to learn a world model, but rather to learn a set of reusable skills that can be composed when solving a given task. These skills are learned to be diverse by iteratively sampling a skill from a skill prior and encouraging a skill-conditioned policy to produce transitions that are predictable, given the sampled skill, and different from those produced by the policy conditioned on a different skill. At test-time, model-predictive control is used to find an optimal sequence of learned skills for solving a target task without any learning on the task. One issue with the proposed method is that the learned skill-conditioned transition model is queried at test-time on states generated by the model itself at previous timesteps, which can be different from the state distribution it was trained on.

Another approach extracts reusable skills from an experience dataset collected across different tasks and recombines them to efficiently solve a downstream task \cite{pertsch2022a}. A variational encoder computes a skill embedding for each trajectory sampled from the dataset and a low-level policy is trained with behavior cloning to decode the embedding into its corresponding actions. State-conditioned skill prior and posterior are trained to match the pretrained skill encoder on behaviors from the experience dataset and task demonstrations, respectively. To solve a downstream task, a high-level policy over skill embeddings is trained with reinforcement learning using an objective that constrains the policy to be close to the skill posterior if the environment state comes from the demonstration data, or to the skill prior otherwise. The approach is shown to outperform prior works that use either demonstrations or task-agnostic experience. However, it makes a strong assumption that the experience dataset contains meaningful, short-horizon behaviors and requires training a separate high-level policy from scratch for every new downstream task.

Our approach shares a common objective with this group of approaches, which is to enable fast adaptation to downstream tasks. However, it stands out by not relying on a world model or an experience dataset.

\subsection{Learning Task-Conditioned Policies}
Lynch et al. \cite{lynch2020learning} propose a method for learning a continuum of robotic tasks from unlabeled play data. The method learns a latent plan distribution space from play sequences by optimizing for reconstruction of play actions while maximizing the similarity between the latent plan distribution of each sequence and that of its combined initial and final states. At test-time, a latent plan is sampled given the current and goal states. It is then fed with the two states to a stochastic policy trained to reconstruct the actions of a play sequence from its corresponding latent plan, initial, and final states. While not requiring expensive expert demonstrations, the proposed method trains the policy on trajectories of play data collected by curious exploration that aims to sufficiently cover the state-action space without regard to the quality of the generated behavior. This leads to a poor policy when the training trajectories are far from the optimal behavior. \par To enable zero-shot generalization to novel tasks, Jang et al. \cite{jang2022bc} propose to condition the policy on information that describes the task such as language instruction or video demonstration. A task embedding is computed from this information and passed into the policy which is supervised with behavior cloning to match the actions in the task demonstrations. The video encoder is constrained to produce an output that is close to the pretrained embedding of the corresponding language description to align the videos more semantically. While the trained policy is found to generalize to unseen tasks, the performance of the video-conditioned policy is lower than the language-conditioned one. The method also requires a predefined task dataset on which the policy is trained at once, and hence the method is not applicable when tasks are presented over time. \par Sodhani et al. \cite{sodhani2021a} found that learning context-based composable representations is an efficient method for sharing information across tasks in multi-task reinforcement learning. In their approach, a natural language task description acts as the context and a mixture of encoders is used to give multiple representations to an input observation. The context determines how to compose the representations by computing soft-attention weights over representations. The weighted sum of representations is concatenated with the context vector and fed to the policy network. Despite improving knowledge transfer, the approach strongly relies on the language description's semantics to extract task-relevant object and skill representations and infer similarity between tasks. It is therefore incompatible with other forms of task description, including visual demonstration.

While our approach, like this group of approaches, conditions the policy on a task description, it is not limited by the quality of task demonstrations or language semantics to extract task-relevant information.

\subsection{Cross-Task Adaptive Regularization}
To accelerate learning new tasks while preserving performance on previously learned tasks, Schwarz et al. \cite{schwarz2018progress} propose to use two neural networks: an active column and a knowledge base. The former is used to learn a new task and is layer-wise connected with the latter to utilize past information. After a task is learned, the active column is distilled into the knowledge base whose parameters are regularized to be close to those adapted to older tasks. The approach has two limitations when used to train a multi-task policy. First, due to the regularization constraint, the knowledge base will learn a policy that tries to achieve average performance on all encountered tasks instead of optimal performance on each individual task. Second, the approach does not address task inference and assumes that changes in task distribution are known to the learner. Hessel et al. \cite{hessel2019multi} suggest that to improve the performance of multi-task reinforcement learning, all tasks should influence the learning updates similarly regardless of the density and scale of their rewards. They propose an actor-critic method that learns multiple tasks in parallel by assigning different environments to different actors. Iteratively, the experience collected by all actors is used to update both a value network with multiple outputs, one for each task, and a task-agnostic policy network used by the actors. The targets in the updates are adaptively normalized by tracking the statistics of the return in each task, causing tasks with different return scales to have similar impact on the learning. A limitation to the proposed method is that the output layer of the value network depends on a predefined number of tasks, rendering it infeasible for learning new tasks over time. Furthermore, parallel training is resource-intensive and impractical for real robots.

Similar to \cite{schwarz2018progress} and \cite{hessel2019multi}, our approach trains a multi-task policy with reinforcement learning, but does not require prior knowledge of task distribution or value network reconfiguration when new tasks are added.

\subsection{Leveraging Task Relationships}
While most approaches to multi-task learning give little consideration to task relationships beyond learning generalizable task features from task examples, a few have shown that exploiting the relationships between tasks leads to fast adaptation to new tasks \cite{oh2017zero,lampinen2020transforming,hafez2021behavior,kalashnikov2021mt}. These approaches mainly differ in how task relationships are learned. For example, Oh et al. \cite{oh2017zero} add an analogy-making objective to encourage the task representation to capture task similarities when optimizing performance over tasks. Kalashnikov et al. \cite{kalashnikov2021mt} propose a distributed multi-task reinforcement learning method in which off-policy data is collected by multiple robots and shared between similar tasks to improve the efficiency of learning each task. This training data is re-balanced between tasks in every training batch when updating the multi-task policy. Besides requiring a predefined discrete set of tasks, that does not allow for continual multi-task learning, the method also requires to manually decide which tasks are semantically similar in order to share data between them. Another approach is to train a meta-mapping function that transforms a learned task representation into another one using a training dataset of task representation pairs, where all paired tasks are systematically related \cite{lampinen2020transforming}. This direct exploitation of systematic relationships has shown better adaptation performance than the indirect way of generalizing through language alone. Instead of relying on prior knowledge in terms of pairs of systematically related tasks \cite{lampinen2020transforming,kalashnikov2021mt} or predefined task analogies \cite{oh2017zero}, a more recent work \cite{hafez2021behavior} learns task relationships unsupervised by continually self-organizing visual demonstrations of tasks so that behaviorally similar tasks are located close to each other. However, the proposed method makes a strong assumption that task demonstrations are perfect and complete, which is restrictive and not often realistic in practice.

The approach described in this paper exploits task relations and learns them in an unsupervised manner from unlabelled task demonstrations, similar to \cite{hafez2021behavior}. The difference is that our approach does not assume completeness of the demonstrations, which makes it more robust and applicable in real-world scenarios where complete demonstrations may not be available.

\section{Technical Approach}\label{sec:approach} 

\begin{figure*}
	\centering
		\includegraphics[width=\textwidth]{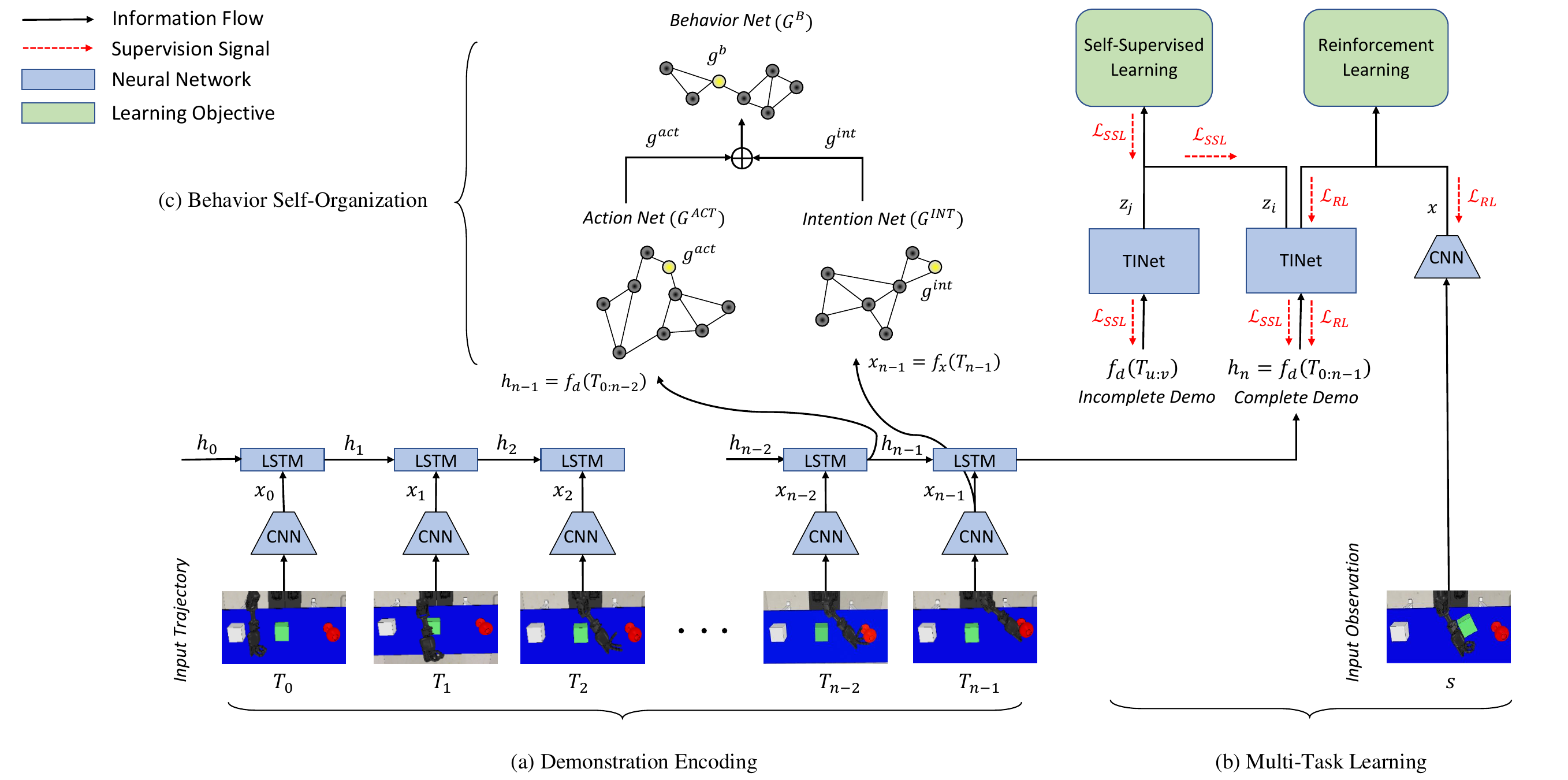}
		\caption{Overview of our proposed task inference architecture for continual multi-task learning. (a) Each input demonstration is encoded with a demonstration encoder $f_d$ by passing image observations in the demonstrated trajectory $T_{0:n-1}$ to a CNN encoder $f_x$ and processing the sequence of CNN-encoded image features $x_{0:n-1}$ with an LSTM. The hidden state $h_n$ after the last feature vector $x_{n-1}$ has been read is used as a latent representation of the demonstration. (b) The TINet is trained with a behavior-matching self-supervised learning objective to map a complete $f_d(T_{0:n-1})$ and an incomplete $f_d(T_{u:v|0\leq u<v\leq n-1})$ version of an input demonstration to a behavior embedding $g^b$ in the self-organizing Behavior Net $G^B$ that best matches the input demonstration. The task representation $z_i$ produced by the TINet is used together with the current observation's feature vector $x = f_x(s)$ as input to a multi-task policy trained with reinforcement learning to optimize performance over tasks. (c) Action and intention embeddings are learned in an unsupervised manner by incrementally self-organizing the movement $f_d(T_{0:n-2})$ and effect $f_x(T_{n-1})$ components of input demonstrations using the growing Action Net $G^{ACT}$ and Intention Net $G^{\textit{INT}}$, respectively. Given an input demonstration, the action $g^{act}$ and intention $g^{int}$ embeddings that best match the demonstrated movement and effect are combined. The behavior embeddings are in turn learned by self-organizing the combined action and intention embeddings with the Behavior Net $G^B$.}
		\label{fig:1}
\end{figure*}

In this section, we present our self-supervised task inference approach for continual multi-task robot learning. We start by describing how action and intention embeddings are learned in an unsupervised manner from unlabeled task demonstrations and used to learn behavior embeddings. Then, the task inference network (TINet) is introduced, which is trained with the proposed behavior-matching self-supervised learning objective. Finally, we show how a multi-task policy can be trained end-to-end with reinforcement learning on top of TINet to optimize performance over tasks.

The aim is to train a multi-task policy with reinforcement learning that can recognize the desired task from an incomplete demonstration and successfully execute the task. At the start of every learning episode, a complete demonstration in the form of a trajectory of $n$ images is randomly sampled and encoded into a vector $h_n$, as shown in Fig. \ref{fig:1}. Similarly, the sequence of the first $n-1$ images is encoded into a vector $h_{n-1}$ and the last image is encoded into a vector $x_{n-1}$. The vectors $h_{n-1}$ and $x_{n-1}$ are used as input to the growing self-organizing networks Action Net and Intention Net, respectively, and the action and intention embeddings $g^{act}$ and $g^{int}$ that best match $h_{n-1}$ and $x_{n-1}$ are identified before the two networks are updated (Sec. \ref{sec:approach-GWR}). The combined action-intention embedding in turn is used as input to the growing self-organizing network Behavior Net, where the behavior embedding $g^b$ that best matches the input is identified and the network is updated. The encoded demonstration $h_n$ is fed to the Task Inference Network (TINet) that outputs the task representation $z_i$. The feature vector of the current environment state $s$ together with $z_i$ are fed to the multi-task policy which outputs the action to take. The TINet is trained with contrastive learning to output the same task representation for the input demonstration (complete demonstration) and for a randomly chosen part of the input demonstration (incomplete demonstration) (Sec. \ref{sec:approach-TINet}). It is also jointly trained to minimize the distance between its output and the behavior embedding $g^b$.

\subsection{Hierarchical Self-Organization of Behaviors}\label{sec:approach-GWR}
\noindent \textbf{Demonstration Encoding.} In our approach, a task demonstration is defined as a trajectory of image observations showing the robot performing a particular behavior to complete the desired task. Any behavior can typically be described by different demonstrations, each being a different trajectory that shows a successful completion of the same task. The image observations in a trajectory are encoded by a Convolutional Neural Network (CNN), and the sequence of the CNN encodings is processed by a recurrent neural network based on the Long Short-Term Memory (LSTM) architecture \cite{lstm1997long} to capture the contextual information in the demonstration. We use the hidden state $h_{n-1}$ of the LSTM after the last CNN encoding $x_{n-1} = f_x(T_{n-1})$ has been read as the latent representation of the entire demonstration, where $n$ is the length of the demonstration. Together, the observation-encoding CNN and context-encoding LSTM define a demonstration encoder $f_{d}$ that takes in a trajectory of observations $T_{0:n-1}$ and outputs a latent representation of the demonstration. Fig. \ref{fig:1}(a) illustrates the demonstration encoding process. Sec. \ref{sec:approach-end-to-endRL} describes how the demonstration encoder is trained. Details on the design choices of the CNN and LSTM networks are given in Sec. \ref{sec4}.

\noindent \textbf{Action and Intention Embeddings.} Each task demonstration is a combination of action and intention, which are the observed movement and effect, respectively. We explicitly leverage this fact and learn two mappings: the first maps from an input space of visually described movements to an embedding space where similar movements are located together; the second maps from an input space of visually described effects to an embedding space where similar effects are located together. These embeddings are called action and intention embeddings, respectively. In our approach, both mappings are learned in an unsupervised manner by incrementally self-organizing the respective input space with a growing self-organizing network. Particularly, we use the Grow When Required (GWR) network \cite{marsland2002self}, which grows when it does not have a close enough match to an input stimulus as opposed to adding nodes at predefined intervals, a criterion often used in other growing networks. This allows adding novel actions and intentions to the respective network once discovered. We refer to the GWR networks used to learn the action and intention embeddings by $G^{ACT}$ and $G^\textit{INT}$, respectively. For each input demonstration, we pass the first $n-1$ observations to the demonstration encoder $f_d$ whose output $f_d(T_{0:n-2})$ is used as input to $G^{ACT}$ and use the CNN-encoded feature vector $x_{n-1}$ of the last observation as input to $G^\textit{INT}$ (see Fig. \ref{fig:1}(c)).

The GWR network is defined by a set of nodes $V$, where each node $i \in V$ is associated with a weight vectors $w_i$, and a set of edges between nodes. At the start of learning, the network has two nodes with weights randomly initialized. In each learning iteration, a new input stimulus $\zeta$ is observed and the following adaptation steps are performed:
\begin{enumerate}
  \item Find the best matching node $c$ and second best matching node $c'$ w.r.t. $\zeta$:
        \begin{equation}
        \label{eq:eq1}
        c = \argmin_{j\in V}  \Vert \zeta - w_j \Vert_2,
        \end{equation}
        \begin{equation}
        \label{eq:eq2}
        c' = \argmin_{j\in {V/\{c\}}}  \Vert \zeta - w_j \Vert_2,
        \end{equation}
        and add an edge between them, if it does not exist, and set its age to 0.
  \item Calculate the activity $a$ of the best matching node based on the Euclidean distance between its weight vector $w_{c}$ and the input $\zeta$:
        \begin{equation}
        \label{eq:eq3}
        a = \exp{(-\Vert \zeta - w_{c} \Vert_2)}.
        \end{equation} 
  \item If the activity $a$ of node $c$ is below a threshold $a_T$ and its habituation (a measure of the node's responsiveness to input stimuli, inversely proportional to the number of times it has been a best match) is below a threshold $h_T$, create a new node $v$ with a weight vector $(w_{c}+\zeta)/2$ and an edge to both $c$ and $c'$ and remove the edge between $c$ and $c'$.
  \item Move the weights of the best matching node $c$ and its neighbors $k$, with which it shares edges, towards $\zeta$:
        \begin{equation}
        \label{eq:eq4}
        \Delta w_{c} = \epsilon_{c} \times h_{c} \times (\zeta - w_{c}),
        \end{equation}
        \begin{equation}
        \label{eq:eq5}
        \Delta w_k = \epsilon_n \times h_k \times (\zeta - w_k),
        \end{equation}
        where $0<\epsilon_n<\epsilon_{c}<1$ and $h_k$ is the habituation value for node $k$.
  \item Decrease the habituation value for the best matching node $c$ and its neighbors $k$:
        \begin{equation}
        \label{eq:eq6}
        h_c = h_0 - \frac{(1-e^{\frac{-\alpha_ct}{\tau_{c}}})}{\alpha_{c}},
        \end{equation}
        \begin{equation}
        \label{eq:eq7}
        h_k = h_0 - \frac{(1-e^{\frac{-\alpha_nt}{\tau_n}})}{\alpha_n},
        \end{equation}
        where $h_0$ is the initial habituation value. $\alpha_{c}$,$\alpha_n$ and $\tau_{c}$,$\tau_{n}$ are constants controlling the habituation curve.
  \item Increment the age of all edges emanating from $c$ by 1. If the age of any edge exceeds a threshold $\kappa$, remove that edge and remove any node with no remaining edges.
\end{enumerate}
 An illustration of the GWR networks $G^{ACT}$ and $G^\textit{INT}$ is shown in Fig. \ref{fig:1}(c).

\noindent \textbf{Behavior Embeddings.}
To learn the behavior embeddings, the combined action-intention embedding space is incrementally self-organized by using a higher-level GWR network $G^B$. During learning, the $G^B$ follows the same adaptation steps of the standard GWR network explained earlier. At each learning iteration, the weight vectors $g^{act}$ and $g^{int}$ of the best matching nodes in the lower-level networks $G^{ACT}$ and $G^\textit{INT}$ w.r.t. an input demonstration are concatenated and fed as input to $G^B$, as shown is Fig. \ref{fig:1}(c). The incremental self-organization of action-intention embeddings allows learning a growing set of behaviors, which is necessary for continual multi-task learning. After the learning of the behavior embeddings, the action $G^{ACT}$, intention $G^\textit{INT}$, and behavior $G^B$ networks can be utilized to map a visual demonstration to the intended behavior behind the demonstration (Algorithm \ref{alg:alg1}). 
\par The two-level hierarchy of embeddings ensures that the learned behavior embeddings capture the action-intention associations and their similarities.

\begin{algorithm}[t]
\caption{$\textsc{Behavior}(T_{0:n-1}) \rightarrow g^b$}
\label{alg:alg1}
\begin{algorithmic}[1]
\small
\REQUIRE Growing self-Organizing networks $G^{ACT}$, $G^\textit{INT}$, and $G^{B}$
\STATE Find the best matching node $c^{act}$ in $G^{ACT}$ w.r.t. $f_d(T_{0:n-2};\theta^{f_d})$
\STATE Compute action embedding $g^{act} \leftarrow w_{c^{act}}$
\STATE Find the best matching node $c^{int}$ in $G^\textit{INT}$ w.r.t. $f_x(T_{n-1};\theta^{f_x})$
\STATE Compute intention embedding $g^{int} \leftarrow w_{c^{int}}$
\STATE Find the best matching node $c^b$ in $G^B$ w.r.t. $g^{act}\oplus g^{int}$
\STATE Compute behavior embedding $g^b \leftarrow w_{c^b}$
\STATE Return $g^b$
\end{algorithmic}
\end{algorithm}

\subsection{Self-Supervised Learning of Task Representations}\label{sec:approach-TINet}
Given the learned behavior embedding space, we aim to train a differentiable model that maps an unlabeled demonstration to a corresponding task representation. In order to do so, we transfer knowledge from the behavior self-organization explained earlier to a task inference neural network, which we call TINet. We construct a behavior-matching self-supervised learning objective to perform the knowledge transfer by training the TINet to map a given demonstration to its nearest behavior embedding in the unsupervisedly learned behavior embedding space, which we use as the target task representation.
\par The input to the TINet is an encoding of a demonstrated trajectory $T_{0:n-1}$ produced by the demonstration encoder $f_d$ and the target output is the weight vector $g^b$ of the best matching node in the behavior net $G^B$ w.r.t. the input demonstration. The TINet model is formally described by:
\begin{equation}
\label{eq:eq8}
z = f_\textit{INF}(f_\textit{d}(T_{0:n-1})),
\end{equation}
where $f_\textit{INF}$ is the task inference function and $z$ is the predicted task representation. We train the TINet to minimize the following behavior-matching loss:
\begin{equation}
\label{eq:eq9}
\mathcal{L}_\textit{BM} =  \Vert f_\textit{INF}\left(  f_\textit{d}(T_{0:n-1}) \right) - g^b    \Vert _{2}^{2}.
\end{equation}
\par To enable task inference from incomplete demonstrations, the TINet is further trained to produce the same output for the original version $T_{0:n-1}^i$ (the complete demonstration) and the temporally cropped version $T_{u:v|0\leq u<v\leq n-1}^j$ (the incomplete demonstration) of an input demonstration (Fig. \ref{fig:1}(b)) in a set of $K$ different demonstrations, where $u$ and $v$ are sampled at random from $[0,n-1]$. This is performed by minimizing the following contrastive loss:
\begin{equation}
\label{eq:eq10}
\mathcal{L}_\textit{C} = -\log \frac{\exp(sim(z_i, z_j)/\tau)}{\sum_{k=0}^{K-1} \mathbbm{1}_{[k \neq i]}\exp(sim(z_i, z_k)/\tau)},
\end{equation}
where $z_i$ and $z_j$ are the task representations of the complete and incomplete versions of the input demonstration, respectively, $\mathbbm{1}_{[k \neq i]}\in \{0,1\}$ is an indicator function, $\tau$ is a temperature hyperparameter, and $sim(\cdot,\cdot)$ is a similarity function. We use cosine similarity $sim(z_i, z_j)= z_i^\top z_j/ (\Vert z_i \Vert_2 \, \Vert z_j \Vert_2)$ between task representations $z_i$ and $z_j$ \cite{chen2020simple}.
This process is shown in Fig. \ref{fig:1}(b). The two TINet blocks are the same network which produces a task representation $z_i$ when the input is the encoded complete demonstration $f_\textit{d}(T_{0:n-1})$ and a task representation $z_j$ when the input is the encoded incomplete demonstration $f_\textit{d}(T_{u:v})$. The contrastive loss in Eq. \ref{eq:eq10} enforces that the task representations for the original (``complete") and cropped (``incomplete") demonstrations $z_i$ and $z_j$, respectively, are close to each other while also preventing the TINet from always producing the same vector on the output by pushing away the task representations of different demonstrations from each other. In other words, minimizing $\mathcal{L}_\textit{C}$ (Eq. \ref{eq:eq10}) means that the complete and incomplete versions of the input demonstration will have nearly identical task representations. Consequently, the robot can now infer the correct task representation when shown only an incomplete demonstration, because the TINet is trained to produce a task representation for the incomplete demonstration that is close to the task representation for the corresponding complete demonstration.

The overall self-supervised learning loss to train a randomly initialized TINet is:
\begin{equation}
\label{eq:eq11}
\mathcal{L}_\textit{SSL} = \mathcal{L}_\textit{BM} + \mathcal{L}_\textit{C}.
\end{equation}
By jointly optimizing for behavior-matching and contrastive prediction, as shown in Eq. \ref{eq:eq11}, the TINet is trained to infer the task at hand from complete or incomplete visual demonstrations without requiring predefined task labels. In addition, distilling the growing behavior net $G^B$ into the TINet (Eq. \ref{eq:eq9}) allows the TINet to continually learn to infer new tasks.

\subsection{End-to-End Continual Multi-Task Learning} \label{sec:approach-end-to-endRL}
Given an unlabeled input demonstration, the learning agent can use the TINet to infer the task it is required to solve. To enable learning a growing set of tasks, we use the task representation $z$ provided by the TINet together with the current environment state $s$ as input to a multi-task policy $\pi$ which is trained with Reinforcement Learning (RL) to optimize performance over tasks. This is done by finding the policy $\pi$ that minimizes the following loss:

\begin{equation}
\label{eq:eq12}
\mathcal{L}_\textit{RL} = - E_\pi \left(    \sum _{t=0}^{\infty} \gamma^t r_t    \right),
\end{equation}
where $t$ is the time step, $r$ is the reward, and $\gamma \in [0,1)$ is a discount factor. Minimizing $\mathcal{L}_\textit{RL}$ corresponds to the standard RL objective of maximizing the expected cumulative discounted reward. The RL algorithm used to train the multi-task policy in the presented work is Deep Deterministic Policy Gradient (DDPG)\cite{lillicrap2015continuous}, but our approach can be paired with any other RL algorithm with minimal changes. DDPG updates the policy by gradient ascent on the action-value ($Q$-)function:
\begin{equation}
\label{eq:eq13}
\theta^\pi = \theta^\pi + \mu \triangledown _{a}Q \left( s,a; \theta^Q \right) \vert _{a= \pi  \left( s_{i} \right) } \triangledown _{ \theta^\pi}\pi \left( s;\theta^\pi \right),
\end{equation}
where $Q(s,a)$ is the expected value of taking action $a$ in state $s$ and following policy $\pi$ thereafter, $\theta^\pi$ and $\theta^Q$ are the policy and $Q$-function parameters, and $\mu$ is the gradient step size. In our implementation, the state feature vector $x = f_x(s;\theta^{f_x})$ and the task representation $z$ are used instead of $s$ as input to $Q$ and $\pi$.

The whole learning architecture (Fig. \ref{fig:1}), including the task inference network TINet, is trained end-to-end. This encourages the task representations from the TINet to capture the task structure. Gradients from $\mathcal{L}_\textit{RL}$ and $\mathcal{L}_\textit{SSL}$ are backpropagated through the TINet, the demonstration encoder $f_d$, and the CNN state encoder $f_x$, optimizing all the networks end-to-end, as shown in Fig. \ref{fig:1}(b). Hence, the final loss for training the multi-task learning agent with our approach is:
\begin{equation}
\label{eq:eq14}
\mathcal{L}_\textit{total} = \mathcal{L}_\textit{SSL} + \mathcal{L}_\textit{RL}.
\end{equation}

The complete Self-Supervised Task Representation Learning (SSTRL) algorithm for continual multi-task RL is given in Algorithm \ref{alg:alg2}. At the beginning of each learning episode, a visual demonstration in the form of a trajectory of frames $T_{0:n-1}$ is presented to the robot (line 5). We then compute the behavior embedding $g^b$ (line 6) of the best matching node in the growing behavior net $G^B$ w.r.t. $T_{0:n-1}$, as shown in Algorithm \ref{alg:alg1}, followed by adjusting the networks $G^{ACT}$, $G^\textit{INT}$, and $G^{B}$(Eq.\ref{eq:eq1}{--}\ref{eq:eq7}). Random temporal cropping is performed on $T_{0:n-1}$ to generate a corresponding incomplete demonstration $T_{u:v}$ (line 8). The indices $u$ and $v$ are uniformly sampled from $[0,n-1]$ such that $0\leq u<v \leq n-1$. We add the complete $T_{0:n-1}$ and incomplete $T_{u:v}$ demonstrations along with the corresponding behavior embedding $g^b$ to the training dataset $\mathcal{D}_{SSL}$ used for minimizing $\mathcal{L}_\textit{SSL}$ (line 9). The encoded demonstration $f_d(T_{0:n-1};\theta^{f_d})$ is passed to the TINet to infer the task representation $z = f_\textit{INF}(f_d(T_{0:n-1};\theta^{f_d});\theta^{f_\textit{INF}})$ (line 10). Actions are generated by the behavioral policy $\pi$ of the chosen RL algorithm $\mathbb{A}$ (line 14). The policy takes as input the state feature vector $x = f_x(s;\theta^{f_x})$ and the inferred task representation $z$ and is parameterized by parameters $\theta^\pi$. $\mathbb{A}$ can be a policy-gradient algorithm, in which case $\theta^\mathbb{A} = \{\theta^\pi\}$, or a value-based algorithm, in which case $\theta^\mathbb{A} = \{\theta^\pi, \theta^Q\}$, where $Q$ is the action-value function. In our implementation, the RL algorithm $\mathbb{A}$ used is DDPG \cite{lillicrap2015continuous} which is value-based and updates the policy by gradient ascent on the $Q$-function (Eq. \ref{eq:eq13}) using experiences sampled from $\mathcal{D}_{RL}$ (line 18). The learning parameters of our task inference architecture are updated to minimize the total loss $\mathcal{L}_\textit{total}$ (Eq. \ref{eq:eq14}).

\begin{algorithm}[t]
\caption{SSTRL algorithm for continual multi-task RL}
\label{alg:alg2}
\begin{algorithmic}[1]
\small
\REQUIRE An RL algorithm $\mathbb{A}$
\REQUIRE 
\parbox[t]{\dimexpr\linewidth-\algorithmicindent*3}{State encoder $f_x$, demonstration encoder $f_d$, task representation encoder $f_\textit{INF}$, and components of $\mathbb{A}$}
\STATE Initialize growing self-organizing networks: $G^{ACT}$, $G^\textit{INT}$, $G^{B}$
\STATE Initialize datasets $\mathcal{D}_{SSL}, \mathcal{D}_{RL} \leftarrow \emptyset$
\STATE Randomly initialize network parameters: $\theta^{f_x}, \theta^{f_d}, \theta^{f_\textit{INF}}, \theta^{\mathbb{A}}$
\FOR {$episode = 1, E$}
    \STATE Sample demonstration $T_{0:n-1}$
    \STATE Compute $g^b = \textsc{Behavior}(T_{0:n-1})$ using Algorithm \ref{alg:alg1}
    \STATE 
        \parbox[t]{\dimexpr\linewidth-\algorithmicindent}{ Adjust $G^{ACT}$, $G^\textit{INT}$, and $G^B$ networks using Eq.\ref{eq:eq1}{--}\ref{eq:eq7} (refer to Sec. \ref{sec:approach-GWR})}
    \STATE $T_{u:v} \leftarrow \textsc{TemporalCrop}(T_{0:n-1})$
    \STATE Insert $(T_{0:n-1},T_{u:v},g^b)$ into $\mathcal{D}_{SSL}$
    \STATE $z \leftarrow f_\textit{INF}(f_d(T_{0:n-1};\theta^{f_d});\theta^{f_\textit{INF}})$
    \STATE Sample initial state $s$
    \WHILE {not terminal}
        \STATE $x \leftarrow f_x(s;\theta^{f_x})$
        \STATE Sample action $a \sim \pi(x,z)$ using $\mathbb{A}$'s behavioral policy
        \STATE Execute $a$ and observe $r$ and $s'$
        \STATE Insert $(s,T_{0:n-1},a,r,s')$ into $\mathcal{D}_{RL}$
        \STATE Update $\theta^{f_x}, \theta^{f_d}, \theta^{f_\textit{INF}}$ using $\mathcal{D}_{SSL}$ and $\mathcal{L}_\textit{SSL}$ in Eq. \ref{eq:eq11}
        \STATE \parbox[t]{\dimexpr\linewidth-\algorithmicindent*2}{Update $\theta^{f_x}, \theta^{f_d}, \theta^{f_\textit{INF}}, \theta^{\mathbb{A}}$ with $\mathbb{A}$ using $\mathcal{D}_{RL}$ and $\mathcal{L}_\textit{RL}$ in Eq. \ref{eq:eq12}}
        \STATE $s \leftarrow s'$
    \ENDWHILE
\ENDFOR
\STATE Return optimized policy $\pi$
\end{algorithmic}
\end{algorithm}

\section{Experimental Results}\label{sec4}
In this section, we empirically evaluate our proposed SSTRL algorithm for continual multi-task RL on learning a fixed as well as a growing set of visuomotor tasks with a humanoid robot and compare it against three multi-task learning baselines: Plan2Explore \cite{sekar2020planning}, Progress\&Compress (P\&C) \cite{schwarz2018progress}, and Behavior-Guided Policy Optimization (BGPO) \cite{hafez2021behavior}. Additionally, we perform ablation experiments to investigate the effect of the different components of our approach on its performance. We also perform one-shot generalization experiments to test the performance of our approach on a set of unseen tasks based on a single visual demonstration. 

\subsection{Experimental Setup}
\label{sec:experimental-setup}

\noindent\textbf{Hyperparameters.} 
The CNN state encoder has three 3$\times$3 convolutions with 32, 64, and 128 channels, respectively. Each convolution is followed by ReLU activation and 2$\times$2 max-pooling. This is followed by two fully connected layers each with 128 units and ReLU activations. The demonstration encoder uses a single-layer LSTM with 256 hidden units and tanh activations. The TINet is a fully connected multi-layer perceptron (MLP) of three layers with 512, 512, and 256 units, respectively. The multi-task policy and $Q$-functions are parameterized by a 2-layer MLP each. The hidden layer is 64-dimensional with ReLU activation. The output layer contains a single unit with linear activation in the $Q$-network and $d$ units with tanh activations in the policy network, where $d$ denotes the dimenstionality of the action space. The training datasets $\mathcal{D}_{SSL}$ and $\mathcal{D}_{RL}$ are stored in memory buffers of sizes $10^5$ and $10^6$, respectively. All networks are trained using the Adam optimizer \cite{kingma2014adam} with learning rate 0.001 and batch size 256. The discount factor $\gamma$ is set to 0.99. We do not use any hyperparameter for balancing the behavior-matching loss and the contrastive loss to simplify the training process. The details on the hyperparameters of the growing Action Net $G^{ACT}$, Intention Net $G^\textit{INT}$, and Behavior Net $G^B$ are given in Appendix \ref{append:1}. Training is done with Tensorflow \cite{abadi2016tensorflow} on a desktop with Intel i5-6500 CPU and a single NVIDIA Geforce GTX 1050 Ti GPU.\par

\begin{figure}[t]
	\centering
		\includegraphics[width=0.75\linewidth]{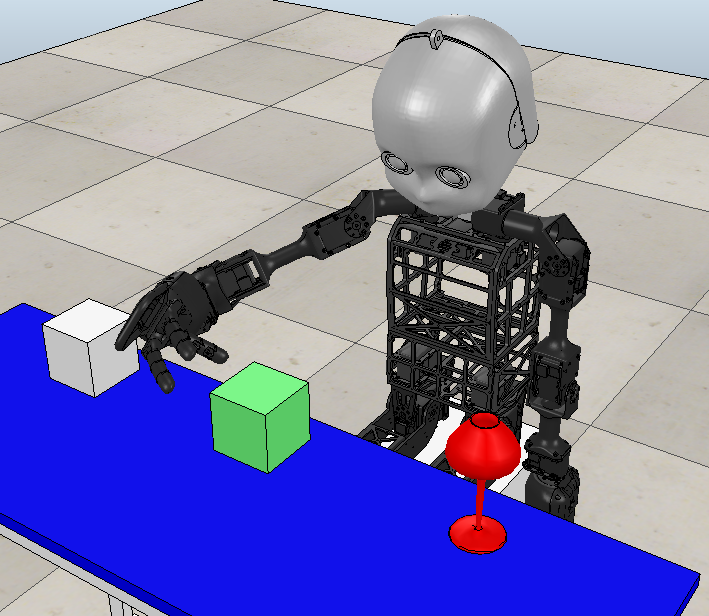}
		\caption{The NICO robot in the simulation environment facing a table with three objects.}
		\label{fig:2}
\end{figure}

\begin{figure}[t]
	\centering
		\includegraphics[width=\linewidth, height=0.6\linewidth]{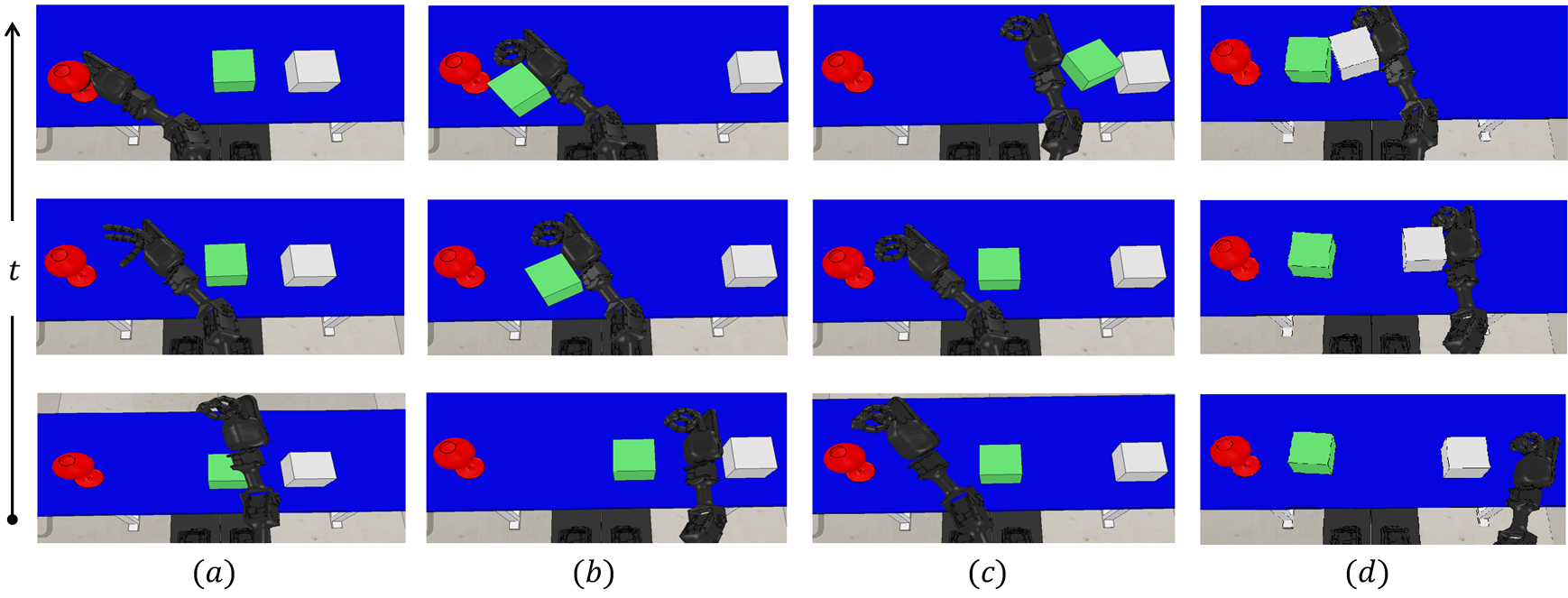}
		\caption{First-person demonstrations of four visuomotor tasks: (a) ``Grasp the red glass", (b) ``Push the green box towards the red glass", (c) ``Push the green box towards the white box", (d) ``Push the white box towards the green box". From bottom to top: RGB frames of initial, intermediate, and terminal configurations.}
		\label{fig:3}
\end{figure}

\noindent\textbf{Robotic setup.} We conduct our experiments on the Neuro-Inspired COmpanion (NICO) robot \cite{kerzel2017nico} using the CoppeliaSim (formerly V-REP) robot simulator \cite{rohmer2013v}. Real-world experiments are described in Sec. \ref{sec:real-world experiments}. Fig. \ref{fig:2} shows the simulated NICO sitting in front of a table on top of which different objects are placed. In all experiments, we consider a motor action controlling four degrees of freedom in the right arm: two joints in the shoulder, one joint in the elbow, and one joint in the hand. The shoulder and elbow joints have an angular range of motion of $\pm{100}$ and $\pm{85}$ degrees respectively. The tendon-operated multi-fingered hand consists of 1 thumb and 2 index fingers with finger joints having an angular range of motion of $0{-}160$ degrees. The input to the robot learning algorithm is a 64$\times$32 RGB image obtained from a vision sensor. Examples of the original output of the vision sensor are shown in Fig. \ref{fig:3}.

\subsection{Multi-Task Learning Evaluation}
In our experiments, we consider the following visuomotor tasks: Grasp the red glass (\textbf{Task-1}), push the green box towards the red glass (\textbf{Task-2}), push the green box towards the white box (\textbf{Task-3}), and push the white box towards the green box (\textbf{Task-4}). We collect 1000 visual demonstrations per task with random initial robot configuration and object positions (see Fig. \ref{fig:3}). The demonstrations have an average length of 30 steps ($\approx 6$s). Due to the demonstrations having variable lengths, we apply zero-padding and masking to them when training the LSTM network of the demonstration encoder. During learning, a task is randomly sampled at the start of each episode which terminates when the task is successfully completed or after a maximum of 50 timesteps. A reward of 1 is given for a successful task completion and 0 otherwise. At the end of each episode, the learned policy is tested by randomly sampling a task and running the policy for 50 timesteps.

\begin{figure}[t]
	\centering
		\includegraphics[width=\linewidth]{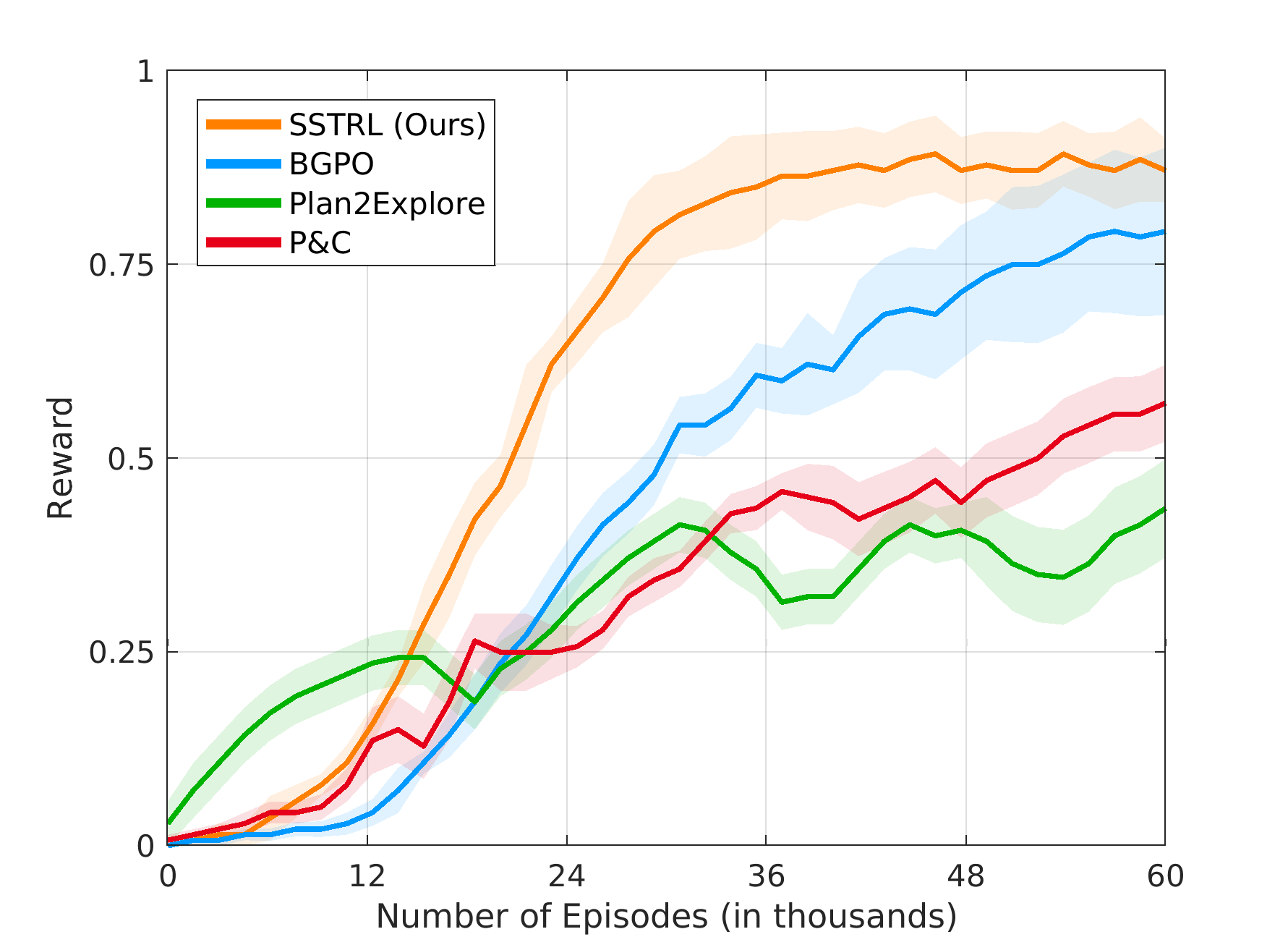}
		\caption{Performance curves of P\&C, Plan2Explore, BGPO, and SSTRL on learning a fixed set of four independent visuomotor control tasks with the NICO robot. Shaded regions represent one standard deviation over 10 random seeds.}
		\label{fig:4}
\end{figure}

\par All learning algorithms use the environment state as input to the policy. Plan2Explore and P\&C assume that changes in task identity are known to the agent, while BGPO and SSTRL perform task inference from unlabeled input demonstrations and use the inferred task representation as additional input to the policy. For implementing Plan2Explore, we train a global world model from task-agnostic experience gathered during learning by an exploration policy trained to maximize the expected novelty over future model states. In each test episode, the policy for the sampled task is trained in imagination using the model and then executed for 50 timesteps. P\&C has two policy networks with layerwise connections between them: the active column and the knowledge base. At each learning episode, the active column is trained on a sampled task and then distilled into the knowledge base which is then evaluated. BGPO and SSTRL receive an unlabeled demonstration from a randomly sampled task at the start of each episode and use it to infer the task representation. The policy conditioned on the inferred representation and environment state is trained for one episode and evaluated on a random task. The implementation details of the baseline algorithms can be found in Appendix \ref{append:2}. We perform our experiments in two settings: (i) fixed-set multi-task learning, where the set of tasks the robot is required to learn is kept fixed throughout the course of learning, and (ii) continual multi-task learning, where the tasks are presented sequentially to the robot.\par Fig. \ref{fig:4} shows the total reward per test episode for each algorithm in the fixed-set multi-task learning setting, averaged over 10 random seeds. As shown in the figure, Plan2Explore performs slightly better than the other algorithms over the first 12K episodes. However, the performance tends to be largely unstable thereafter, with the average reward staying under 0.5 (i.e., below 50\% success rate). This is likely the result of the world model being trained on parts of the environment that are not relevant to the task at test time. In contrast, the performance of P\&C continues to improve with more training but reaches only an average reward of 0.57 by the end of learning. Since the knowledge base parameters in P\&C are restricted to be close to their previously trained values when the policy learned by the active column is distilled into the knowledge base, the multi-task policy of the knowledge base can only generalize slowly. Consequently, its performance on a given task heavily depends on how similar that task is to the recently learned one, which may explain the slow increase in the observed average reward over time. Instead of mitigating interference among tasks by regularizing the update to the multi-task policy parameters, which still leads to interference since tasks are learned in a joint parameter space, BGPO and SSTRL avoid interference in the first place by conditioning the policy on a task representation which is learned in a space different than that of the policy parameters. While BGPO and SSTRL show a better final performance, achieving an average reward of over 0.75 at the end of learning, SSTRL has a more stable performance and faster convergence than BGPO.
\par We also evaluate the performance of the trained policy of each algorithm on the individual tasks. This includes a comparison to a single-task policy optimization, where a separate policy network is trained with DDPG \cite{lillicrap2015continuous} on each task individually (see Appendix \ref{append:2} for implementation details). The trained policy attempts each task 100 times. The success rate is given in Table \ref{table:table-indiv-tasks}, with SSTRL achieving the highest success rate in 3 out of 4 tasks. The results suggest that multi-task learning with SSTRL not only allows the policy to accomplish a number of different tasks, but also to improve its performance on each individual task via sharing policy and task representations.

\begin{table} [t]
\caption{Performance Comparison of the Algorithms on Individual Tasks. The Reported Numbers are Success Rates Over 100 Trials.}
\centering
\begin{tabular}{l c c c c} 
 \toprule
 \textbf{Algorithm} & \textbf{Task-1} & \textbf{Task-2} & \textbf{Task-3} & \textbf{Task-4}\\
 \midrule
 Single-task policy optimization & 53\% & 72\% & 66\% & 60\%\\
 P\&C \cite{schwarz2018progress} &                            23\% & 55\% & 31\% & 18\%\\
 Plan2Explore\cite{sekar2020planning} &                    35\% & 43\% & 29\% & 25\%\\
 BGPO  \cite{hafez2021behavior} &                           54\% & 82\% & 68\% & \textbf{71\%}\\
 SSTRL (Ours) &                           \textbf{67\%} & \textbf{86\%} & \textbf{79\%} & 68\%\\
\bottomrule
\end{tabular}
\label{table:table-indiv-tasks}
\end{table}

\begin{figure}[t]
	\centering
		\includegraphics[width=\linewidth]{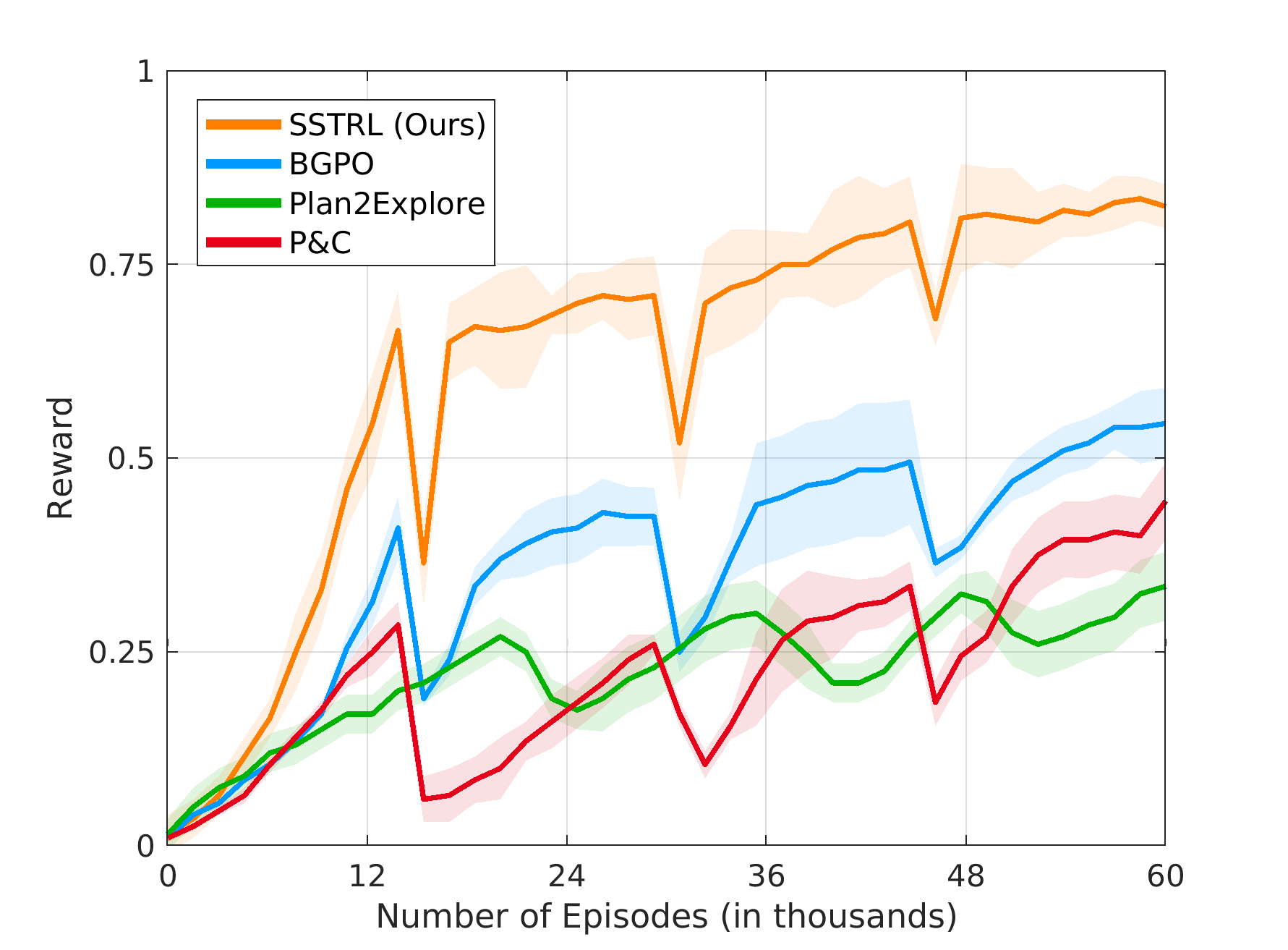}
		\caption{Performance curves of P\&C, Plan2Explore, BGPO, and SSTRL in the continual multi-task learning setting. Shaded regions represent one standard deviation over 10 random seeds.}
		\label{fig:5}
\end{figure}
\par In the continual learning setting, the learning starts with \textbf{Task-1}. \textbf{Task-2}, \textbf{Task-3}, and \textbf{Task-4} are presented after 15K, 30K, and 45K episodes respectively. At test time, each algorithm is evaluated on a task randomly sampled from the presented tasks. We plot the total reward per test episode, averaged over 10 random seeds, in Fig. \ref{fig:5}. A drop in performance can be observed for all algorithms after each new task is introduced. Changing tasks has less direct effect on Plan2Explore as it follows task-agnostic exploration policy during learning. However, the policy learned offline at test time relies on a world model that may have been trained on task-irrelevant data. Thus, the policy performs poorly on tasks for which the model is not sufficiently trained, particularly when such tasks are visited for several episodes before new tasks are presented, as it is the case in the continual learning setting. Similarly, when P\&C encounters an unseen task, the knowledge base's policy typically requires longer training before it adapts to that task, because the active column's policy, which is distilled into the knowledge base after every episode, will likely fail to quickly solve the unseen task. This leads to a considerably small improvement in performance during the intervals between the tasks, as shown in Fig. \ref{fig:5}. \par BGPO and SSTRL, on the other hand, are originally more robust to learning instability caused by the sequential presentation of tasks due to their self-organizing networks that grow when they cannot find a behavior embedding that closely matches an input demonstration. By using a task representation based on a growing behavior network as input to the policy, they can generalize faster to new tasks and maintain the performance on old tasks while learning new ones since different tasks have different representations on which the policy is conditioned. Compared to BGPO that reached an average reward of only 0.55 after 60K episodes, SSTRL was able to converge to an average reward of over 0.8. This empirically shows the advantage of learning separate action and intention embeddings and the advantage of the TINet that learns a generalizable mapping from demonstrations to task representations end-to-end, which is particularly useful in the continual learning setting.

\begin{figure}[t]
	\centering
		\includegraphics[width=\linewidth]{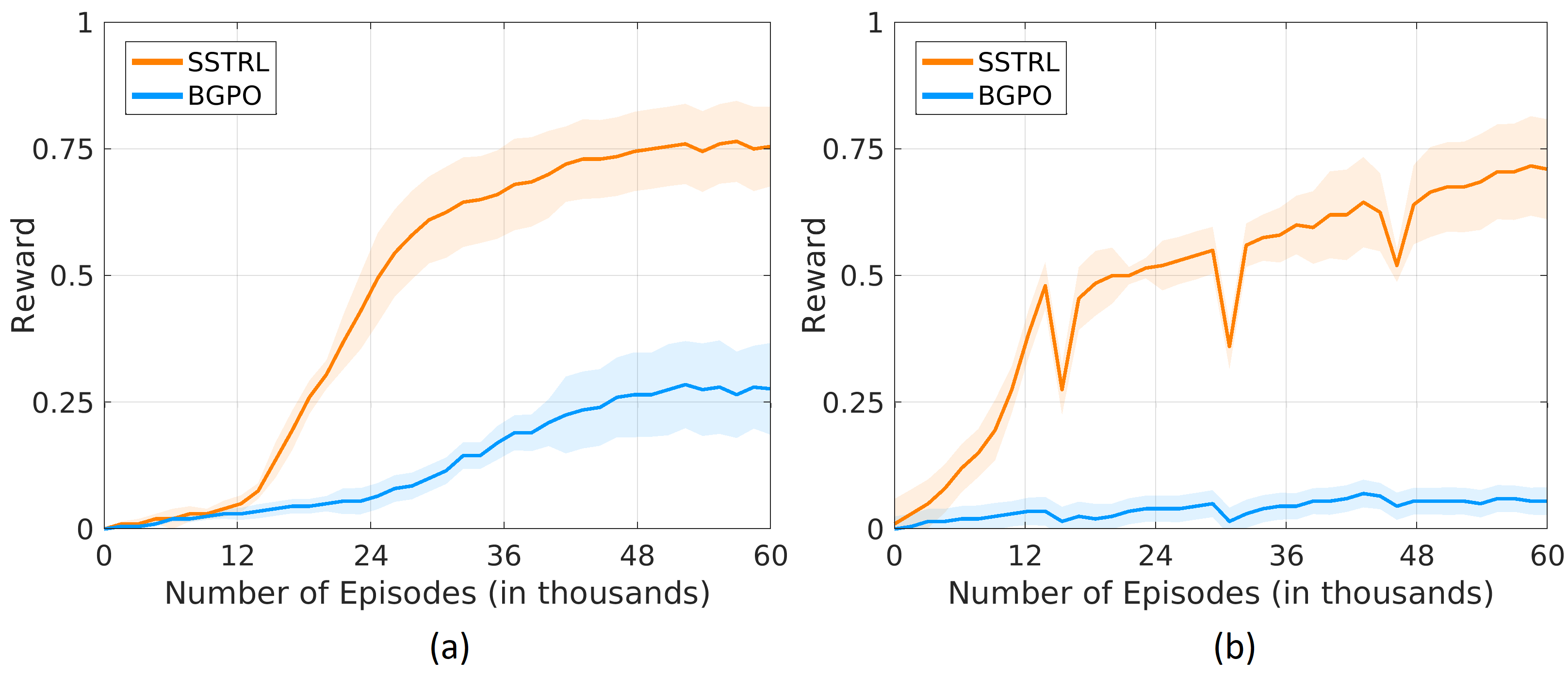}
		\caption{Performance curves of BGPO and SSTRL on incomplete demonstrations in two multi-task learning settings: (a) fixed-set and (b) continual. Shaded regions represent one standard deviation over 10 random seeds.}
		\label{fig:6}
\end{figure}

\subsection{Performance Evaluation on Incomplete Demonstrations}
In this experiment, we aim to compare the performance of the multi-task policy trained with BGPO and SSTRL when incomplete demonstrations are used as input. We make two comparisons in the fixed-set and continual learning settings. At each test episode, we randomly sample 10 unlabeled demonstrations and apply random temporal cropping to each demonstration. We then run the trained policy 10 times each using a different temporally cropped demonstration as input. The results are shown in Fig. \ref{fig:6}. In the fixed-set setting, BGPO reaches an average reward of around 0.25, while SSTRL appears to achieve three times more average reward by the end of learning. The difference in performance between the two algorithms is more pronounced in the continual learning setting. As opposed to BGPO which fails to make any progress in maximizing the obtained reward over the entire learning process, SSTRL is able to continue to improve its performance after every new task is presented. This demonstrates the effectiveness of training the TINet to learn a joint representation of original and cropped versions of unlabeled demonstrations in SSTRL, enabling task inference from incomplete demonstrations and improving the performance of the multi-task policy on tasks inferred form such demonstrations.

\subsection{Ablation Study}
\begin{figure}[t]
	\centering
		\includegraphics[width=\linewidth]{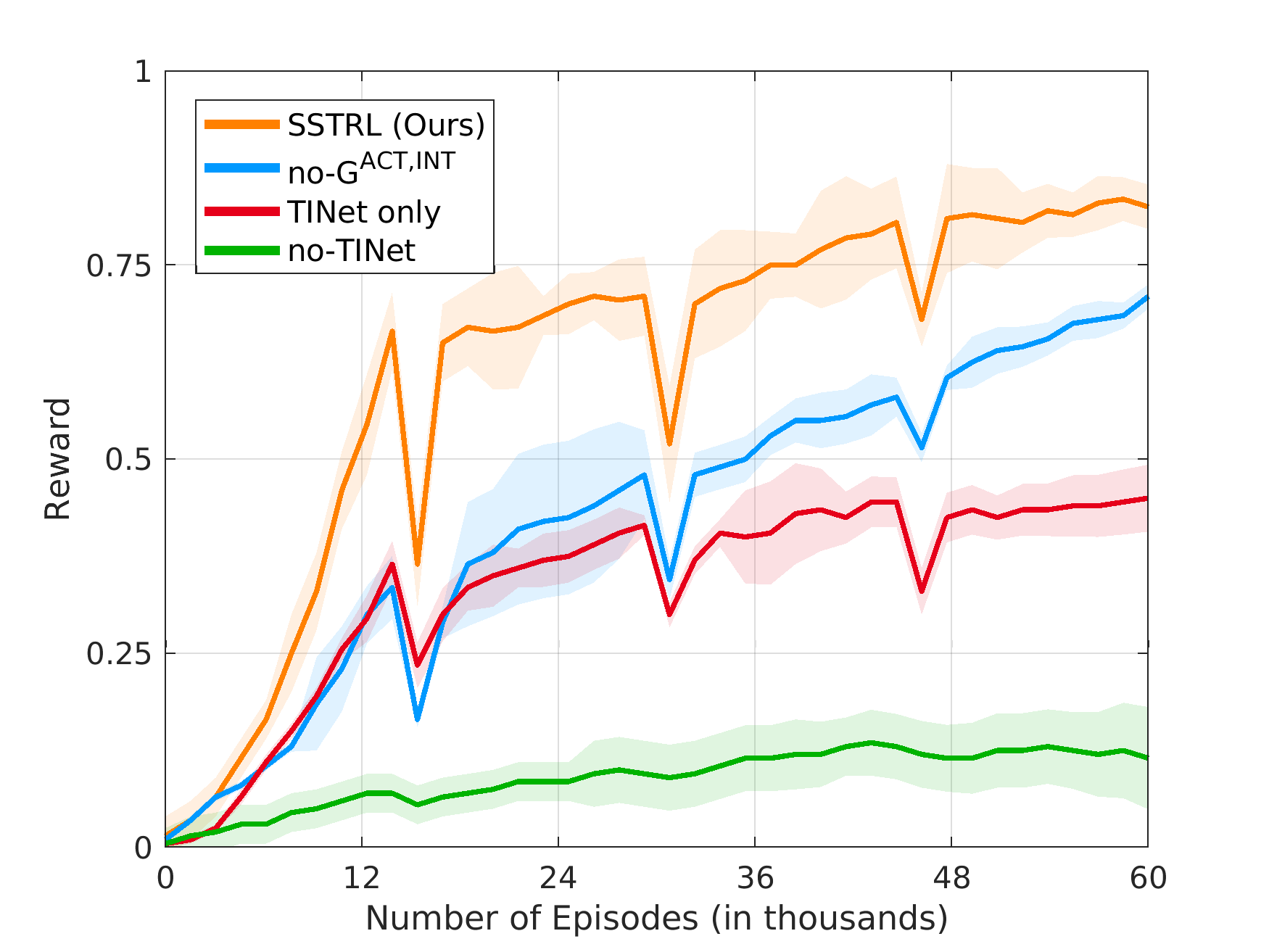}
		\caption{Performance curves of SSTRL with all of its components and after removing different components in the continual multi-task learning setting. Shaded regions represent one standard deviation over 10 random seeds.}
		\label{fig:7}
\end{figure}

We perform an ablation study to investigate the contribution of each component of our proposed approach to the overall performance in the continual learning setting and plot the results in Fig. \ref{fig:7}. When the action and intention networks are removed ("\textit{no-$G^\textit{ACT,INT}$}"), the behavior network $G^B$ is forced to learn the behavior embeddings directly from demonstrations. This slows convergence as the robot lacks the information the action-intention associations offer to facilitate task inference. Removing the hierarchical self-organization architecture altogether and keeping the TINet ("\textit{TINet only}") causes the multi-task policy to converge to a lower average reward than when self-organization of behaviors is enabled. Since in this case the TINet cannot be trained to map demonstrations to best matching behaviors, the algorithm will likely fail to infer the intended behavior behind each demonstration, thus preventing the policy from achieving high performance across all tasks. If the TINet is removed ("\textit{no-TINet}"), the policy exhibits poor performance, with the average reward staying under 0.25 until the end of learning. The success rate per task for each of the considered Configurations is given in
Table \ref{table:table-ablation}.
\par 

These results suggest that behavior self-organization is essential for task inference in SSTRL and that, without the behavior-matching loss, the TINet can only reach an average performance, indicating that training the TINet with the contrastive loss is sufficient for improving the performance compared to the case when the TINet is removed. Additionally, unsupervised learning of action and intention embeddings with $G^\textit{ACT}$ and $G^\textit{INT}$ networks and using the combined action-intention embeddings as input to the $G^B$ network allow behavior self-organization to significantly improve learning speed and final performance compared to directly using the encoded demonstrations as input to $G^B$.

\begin{table} [t]
\caption{Success Rate per Task for Each Ablation Configuration Over the Test Episodes.}
\centering
\begin{tabular}{l c c c c} 
 \toprule
 \textbf{Algorithm} & \textbf{Task-1} & \textbf{Task-2} & \textbf{Task-3} & \textbf{Task-4}\\
 \midrule
no-$G^\textit{ACT,INT}$ &         65.33\% & 61.12\% & 45.80\% & 33.00\%\\
TINet only  &                    42.35\% & 37.19\% & 30.51\% & 17.79\%\\
no-TINet  &                       11.98\% & 10.33\% & 5.56\% & 4.95\%\\
SSTRL (Ours)  &              \textbf{74.32\%} & \textbf{71.00\%} & \textbf{68.76\%} & \textbf{56.20\%}\\
\bottomrule
\end{tabular}
\label{table:table-ablation}
\end{table}

\subsection{One-Shot Task Generalization}
\begin{figure}[t]
	\centering
		\includegraphics[width=\linewidth]{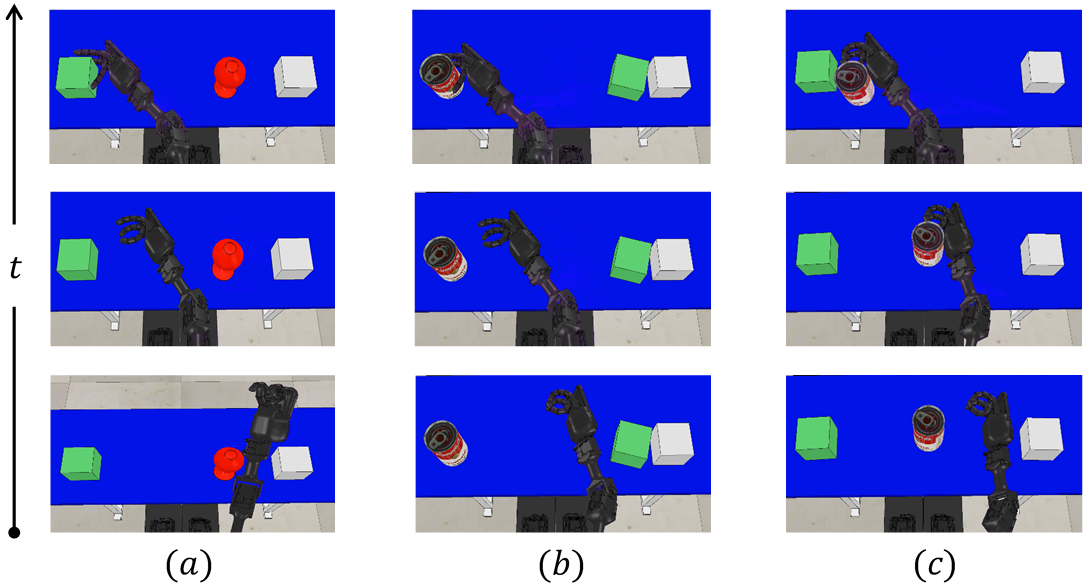}
		\caption{Example demonstrations of three held-out visuomotor tasks: (a) ``Grasp the green box", (b) ``Grasp the can", (c) ``Push the can towards the green box". From bottom to top: RGB frames of initial, intermediate, and terminal configurations.}
		\label{fig:8}
\end{figure}

We evaluate the multi-task policy trained in the continual learning setting with BGPO and SSTRL (the best-performing policy out of 10 training runs) on a held-out set of three unseen tasks: Grasp the green box (\textbf{Task-5}), grasp the can (\textbf{Task-6}), and push the can towards the green box (\textbf{Task-7}). For each unseen task, we provide the trained policy with one visual demonstration of successful task execution as input. Fig. \ref{fig:8} shows an example demonstration for each task. We perform 100 test trials with 100 different demonstrations per unseen task. The success rate for the unseen tasks is given in Table \ref{table:table-one-shot}.

The held-out tasks are chosen such that they include combinations of action-object pairs (\textbf{Task-5}, \textbf{Task-6}) and object-object pairs (\textbf{Task-7}) that were not seen during training. As shown in Table \ref{table:table-one-shot}, SSTRL achieves a non-zero success rate in all the held-out tasks. We find that \textbf{Task-6} is particularly challenging. We believe this is because it not only involves a novel object (the can), but also because the policy has learned to \textit{grasp} only a single object (the red glass) during training whereas it has learned to \textit{push} different objects. Nevertheless, the multi-task policy trained with SSTRL appears to capture the intention of the observed demonstration despite having never seen any demonstration of the related task. We demonstrate the one-shot generalization performance of our approach on the held-out tasks in the accompanying video (\url{https://youtu.be/77cP8ciHcII}). Analyzing the trajectories generated by the policy, we observe that while the robot does not exactly complete the task in the unsuccessful trials, it moves towards the target object, and in many cases the robot does come fairly close to the target object but fails only to close the fingers correctly on the object or to place the pushed object right next to the target object. This clearly indicates that the inferred task representation is informative of the task at hand.

\begin{table}[t]
\caption{One-Shot Generalization Performance on Held-Out Tasks. The Reported Numbers are Success Rates Over 100 Trials.}
\centering
\begin{tabular}{l c c c} 
 \toprule
 \textbf{Algorithm} & \textbf{Task-5} & \textbf{Task-6} & \textbf{Task-7} \\
 \midrule
 BGPO  \cite{hafez2021behavior} &  22\%          & 0\%          & 0\%\\
 SSTRL (Ours) &  \textbf{55\%} & \textbf{35\%} & \textbf{61\%} \\
\bottomrule
\end{tabular}
\label{table:table-one-shot}
\end{table}

\subsection{Performance Evaluation in the Real World}
\label{sec:real-world experiments}
We compare the performance of the multi-task policy trained under the continual learning setting in simulation on the real NICO robot. For each algorithm, we take the best-performing policy out of 10 simulation-based training runs and perform 20 test episodes per task on the real robot, each with different object positions and initial robot configuration. For BGPO and SSTRL, we use a random simulation-based demonstration as input to the policy.

\begin{figure}[t]
	\centering
		\includegraphics[width=.85\linewidth]{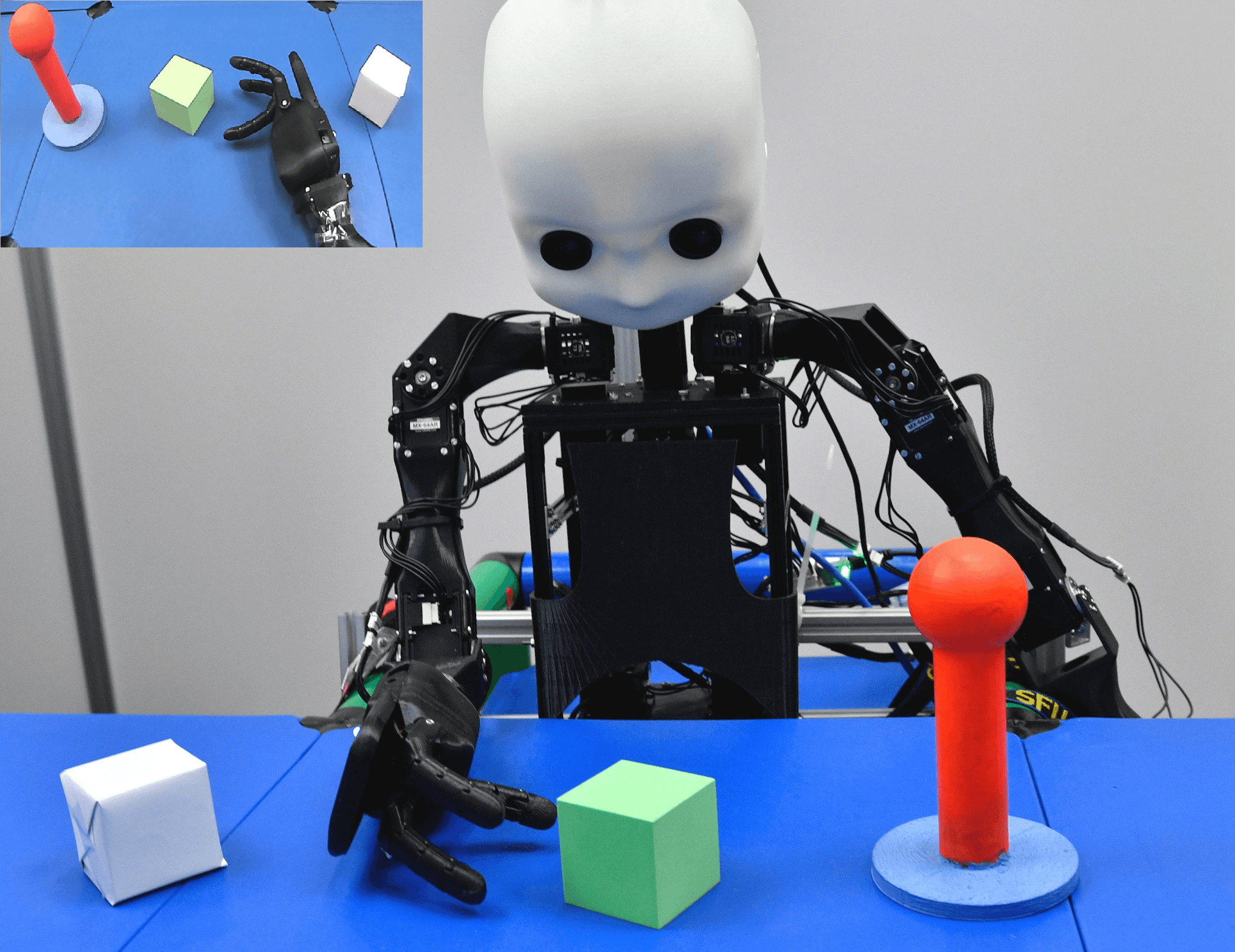}
		\caption{Exocentric and egocentric (inset) views of the real-world experimental setup for multi-task learning from the perspective of the NICO robot.}
		\label{fig:9}
\end{figure}

The simulation uses a URDF model of the NICO robot and, thus, there is no difference between the real NICO and the simulated NICO. The height and color of the table and of the robot's seat are identical in the simulation and the real environment, which facilitates a direct transfer of the robot's arm pose and the trained policy. The objects on the table are slightly different in geometry to enable more stable manipulation but have the same colors as in simulation. Fig. \ref{fig:9} shows the experimental setup with the real NICO robot. We do not perform any additional training or fine-tuning of the learning architecture and deploy it directly on the real NICO.
We report the success rate per task for each algorithm in Table \ref{table:table-real}. Example runs of the multi-task policy learned using SSTRL on the real robot are shown in the accompanying video (\url{https://youtu.be/77cP8ciHcII}).
\begin{table} [t]
\caption{Performance Evaluation in the Real World. The Reported Numbers are Success Rates Over 20 Test Episodes.}
\centering
\begin{tabular}{l c c c c} 
 \toprule
 \textbf{Algorithm} & \textbf{Task-1} & \textbf{Task-2} & \textbf{Task-3} & \textbf{Task-4}\\
 \midrule
 P\&C \cite{schwarz2018progress} &                            15\% & 25\% & 15\% & 20\%\\
 Plan2Explore \cite{sekar2020planning} &                    10\% & 30\% & 20\% & 25\%\\
 BGPO  \cite{hafez2021behavior} &                           35\% & \textbf{75\%} & 50\% & 50\%\\
 SSTRL (Ours) &                           \textbf{60\%} & \textbf{75\%} & \textbf{65\%} & \textbf{70\%}\\
\bottomrule
\end{tabular}
\label{table:table-real}
\end{table}

\section{Conclusion}
In this paper, we presented a novel self-supervised task inference approach for continual robot learning. Our approach uses a two-level hierarchical self-organization architecture to learn task-descriptive behavior embeddings from unlabeled vision-based demonstrations. In the lower level, action and intention embeddings are incrementally learned from self-organization of the observed movement and effect parts of each demonstration. The self-organization of the combined action-intention embeddings constructs a higher-level behavior embedding space. A novel behavior-matching self-supervised learning objective is used to train a task inference network, which we call TINet, to map complete and incomplete versions of an unlabeled demonstration to a best matching behavior in the learned behavior embedding space, which we use as the target task representation. A multi-task policy is built on top of the TINet and trained with reinforcement learning to optimize performance over tasks. The whole learning architecture, including the TINet, is trained end-to-end, encouraging the inferred task representation to capture the structure of the task at hand. We evaluate our approach in the fixed-set and continual multi-task learning settings and compare it to different multi-task learning baselines. The results show that our approach outperforms the other baselines, with the difference being more pronounced in the challenging continual learning setting. Our approach also achieves higher task performance than the compared demonstration-based baseline on incomplete input demonstrations. Additionally, the results from one-shot task generalization experiments clearly demonstrate the ability of our approach to read the intention behind a task demonstration and generate a meaningful action trajectory to complete the task without any learning on the task.
\par In contrast to previous multi-task learning approaches, our approach is constant in the number of policy parameters, makes no assumptions about task distribution, and while maintaining performance on previously learned tasks, avoids learning interference among tasks as it accelerates learning progress on new tasks. The ability to infer the intended behavior behind a visual demonstration rather than copying and memorizing the observed actions allows our approach to complete the tasks more efficiently than the provided demonstrations, especially when the input demonstration is imperfect or incomplete. It is also worth mentioning that the policy trained with our approach performs the desired task even when the initial environment state, including robot configuration, is different between the demonstration and test-time settings, as shown in the accompanying video (\url{https://youtu.be/77cP8ciHcII}).
\par One limitation of our approach is that it requires additional computational time for constructing and adapting the growing self-organizing networks necessary for task inference. However, this happens only once every learning episode and never at test time, where only the TINet is queried for a task representation without any search in the graph of the growing networks. Besides, this computational complexity scales only linearly with the number of demonstrations. In its current form, our approach uses first-person demonstrations. One exciting direction for future work is to adapt our approach to address task inference when the morphology of the demonstrator and the robot are different, which is the case when using third-person demonstrations from a human teacher. Another direction for future work is to train the approach on multimodal demonstrations using vision-and-language task descriptions.

\appendices
\section{Hyperparamters of the Growing Self-Organizing Networks} \label{append:1}
Here we give details on the hyperparameters of the growing Action Net $G^{ACT}$, Intention Net $G^\textit{INT}$, and Behavior Net $G^B$ in Tables \ref{table:table1}, \ref{table:table2}, and \ref{table:table3}, respectively.

\begin{table} [H]
\caption{The Hyperparameters of $G^{ACT}$ Net Used in Our Experiments.}
\centering
\begin{tabular}{l c} 
 \toprule
 Hyperparameter & Value\\
 \midrule
 Activity Threshold & $a_T=0.7$\\
 Habituation Threshold & $h_T=0.2$\\
 Learning Rates & $\epsilon_c=0.1, \epsilon_n=0.05$\\
 Initial Habituation & $h_0=1$\\
 Habituation Curve & $\alpha_c=\alpha_n=1.05, \tau_c=0.5, \tau_n=2$\\
 Maximum Age Threshold & $\kappa=80$\\
\bottomrule
\end{tabular}
\label{table:table1}
\end{table}

\begin{table} [H]
\caption{The Hyperparameters of $G^\textit{INT}$ Net Used in Our Experiments.}
\centering
\begin{tabular}{l c} 
 \toprule
 Hyperparameter & Value\\
 \midrule
 Activity Threshold & $a_T=0.9$\\
 Habituation Threshold & $h_T=0.3$\\
 Learning Rates & $\epsilon_c=0.1, \epsilon_n=0.01$\\
 Initial Habituation & $h_0=1$\\
 Habituation Curve & $\alpha_c=\alpha_n=1.05, \tau_c=1, \tau_n=2.7$\\
 Maximum Age Threshold & $\kappa=100$\\
\bottomrule
\end{tabular}
\label{table:table2}
\end{table}

\begin{table} [H]
\caption{The Hyperparameters of $G^B$ Net Used in Our Experiments.}
\centering
\begin{tabular}{l c} 
 \toprule
 Hyperparameter & Value\\
 \midrule
 Activity Threshold & $a_T=0.8$\\
 Habituation Threshold & $h_T=0.15$\\
 Learning Rates & $\epsilon_c=0.1, \epsilon_n=0.01$\\
 Initial Habituation & $h_0=1$\\
 Habituation Curve & $\alpha_c=\alpha_n=1.05, \tau_c=3.3, \tau_n=14.3$\\
 Maximum Age Threshold & $\kappa=90$\\
\bottomrule
\end{tabular}
\label{table:table3}
\end{table}

\section{Baseline Details} \label{append:2}
\noindent\textbf{Plan2Explore.} 
The convolutional image encoder and decoder networks are the same from \cite{ha2018world}. We use an ensemble of 5 transition models whose prediction disagreement is used to derive the intrinsic reward for the task-agnostic exploration, with each model implemented as a 2 hidden-layer MLP which takes the recurrent state of the recurrent state space model (RSSM) \cite{hafner2019learning} and the action as input and predicts 1024-dimensional image encoder features. The reward prediction, state-value, and policy functions are parameterized by a 3 hidden-layer MLP each, with 200 ReLU units in each layer, and trained using Adam \cite{kingma2014adam} with a learning rate of $10^{-5}$. The imagination horizon is 15.

\noindent\textbf{P\&C.}
The policy and value function of the active column and knowledge base share a convolutional encoder of two Conv layers with 16 6$\times$6 and 32 3$\times$3 filters and ReLU activations, followed by a fully connected layer with 256 units and ReLU activations. A separate, fully connected output layer with linear activation is used for each of the policy and value function. The networks are trained using Adam \cite{kingma2014adam} and with a learning rate of 0.003. The forgetting coefficient and Fischer regularisation strength are set to 0.95 and 25, respectively.

\noindent\textbf{BGPO.} 
A variational autoencoder (VAE) \cite{kingma2014auto} is used to learn the state representation. The VAE encoder consists of three Conv layers with 32, 64, and 128 3$\times$3 filters, each followed by ReLU activation and 2$\times$2 max-pooling, and two fully connected layers of 128 linear units outputting the mean and standard deviation of a diagonal Gaussian from which state representations are sampled. The decoder mirrors the encoder, but uses a sigmoid activation in the output layer. The inverse model used is a 1-layer MLP with tanh activation. Each demonstration is a sequence of VAE-encoded states. An LSTM autoencoder with 64 hidden units in the encoder and decoder LSTMs is used to encode the demonstrations. We use the same hyperparameters that were used in \cite{hafez2021behavior} for the GWR-B model. DDPG is used as the base RL algorithm. The policy and action-value function are parameterized by an MLP with a hidden layer of 64 ReLU units and output layer of one linear unit in the action-value function and 4 tanh units in the policy. The VAE, inverse model and LSTM autoencoder are pretrained on 4000 task demonstrations and then fixed during policy learning. The networks are trained using Adam \cite{kingma2014adam} with learning rate 0.001.

\noindent\textbf{Single-task policy optimization.}
In both the single-task $Q$- and policy networks, a CNN state encoder (described in Sec. \ref{sec:experimental-setup}) is used, followed by one fully-connected hidden layer with 64 ReLU units. The output layer in the policy network is a tanh layer and in the $Q$-network is a linear layer. The actions are included after the CNN encoder in the $Q$-network. The networks are trained with DDPG \cite{lillicrap2015continuous} using Adam \cite{kingma2014adam} with learning rate 0.001 and batch size 256.

\section*{Acknowledgement}
We gratefully acknowledge support from the German Research Foundation DFG under project CML (TRR 169). We thank Erik Strahl for his technical support with the real-world experimental setup.

\bibliographystyle{IEEEtran}
\bibliography{bibliography}
\end{document}